%% file: 00-main.tex
\documentclass[conference]{IEEEtran}

\usepackage{multirow}
\usepackage{makecell}
\usepackage{graphicx}
\usepackage{xcolor}
\usepackage{amsmath,amssymb,amsfonts}
\usepackage{booktabs}
\usepackage[caption=false,font=footnotesize]{subfig}

\begin{document}

\title{Multi-Resolution End-to-End Deep Neural Network for Optimizing Latency-Accuracy Tradeoff in Autonomous Driving}

\author{
\IEEEauthorblockN{Qitao Weng and Heechul Yun}
\IEEEauthorblockA{
University of Kansas\\
Lawrence, KS, USA\\
\{wengqt, heechul.yun\}@ku.edu
}
}

%%
%% By default, the full list of authors will be used in the page
%% headers. Often, this list is too long, and will overlap
%% other information printed in the page headers. This command allows
%% the author to define a more concise list
%% of authors' names for this purpose.
% \renewcommand{\shortauthors}{Trovato et al.}

\maketitle

%%
%% The abstract is a short summary of the work to be presented in the
%% article.
\begin{abstract}
Latency–accuracy tradeoffs are fundamental in real-time applications of deep neural networks (DNNs) for cyber-physical systems. In autonomous driving, in particular, safety depends on both prediction quality and the end-to-end delay from sensing to actuation. We observe that (1) when latency is accounted for, the latency-optimal network configuration varies with scene context and compute availability; and (2) a single fixed-resolution model becomes suboptimal as conditions change.

We present a multi-resolution, end-to-end deep neural network for the CARLA urban driving challenge using monocular camera input. Our approach employs a convolutional neural network (CNN) that supports multiple input resolutions through per-resolution batch normalization, enabling runtime selection of an ideal input scale under a latency budget, as well as resolution retargeting, which allows multi-resolution training without access to the original training dataset.

We implement and evaluate our multi-resolution end-to-end CNN in CARLA to explore the latency–safety frontier. Results show consistent improvements in per-route safety metrics—lane invasions, red-light infractions, and collisions—relative to fixed-resolution baselines.
\end{abstract}

\input{01-intro}
\input{02-background}

\input{03-multires}

\input{04-evaluation}

\input{05-related}
\input{06-conclusion}
\input{07-ack}

%% The acknowledgments section is defined using the "acks" environment
%% (and NOT an unnumbered section). This ensures the proper
%% identification of the section in the article metadata, and the
%% consistent spelling of the heading.
% \begin{acks}
% To Robert, for the bagels and explaining CMYK and color spaces.
% \end{acks}

%%
%% The next two lines define the bibliography style to be used, and
%% the bibliography file.
\bibliographystyle{IEEEtran}
\bibliography{reference}

\end{document}

%% file: 01-intro.tex
\section{Introduction}
End-to-end (E2E) deep-learning–based autonomous driving establishes a cyber-physical control loop in which a deep neural network (DNN), operating on sensor data, directly commands a physical vehicle via actuators. Safety in this loop depends jointly on prediction quality and end-to-end latency: any delay between sensing and actuation causes commands to operate on stale information. In urban autonomous driving settings, perception demands vary substantially along the route: lane-following segments tolerate coarser spatial detail, while intersections require reliable detection of small, safety-critical cues such as traffic lights and cross traffic. Because convolutional inference cost grows with input resolution, accuracy and latency are inherently coupled, motivating approaches that explicitly reason about both.

DNN-based E2E policies are increasingly adopted in many cyber-physical systems (CPS) because they directly optimize the final control objective—jointly tuning perception, prediction, and control—rather than relying on human-defined intermediate targets in modular stacks~\cite{bojarski2016end,codevilla2018end}. Compared with modular pipelines, E2E models offer a unified trainable architecture, align all intermediate representations with the task, improve computational efficiency through shared backbones, and scale effectively with additional data and compute~\cite{chen2024end}.

Prior E2E driving policies~\cite{pomerleau1988alvinn,bojarski2016end} typically fix input resolution and compute, optimizing accuracy while leaving loop-level latency implicit. Some recent studies investigate the latency and accuracy tradeoffs in E2E systems~\cite{khalil2024plm,bechtel2022deeppicarmicro}. 
%including DeepPicarMicro~\cite{bechtel2022deeppicarmicro}, which analyzes accuracy–latency tradeoffs for lane-following tasks on embedded platforms through architectural choices, and 
%PLM-Net~\cite{khalil2024plm}, which mitigates the information staleness by combining action predictions aligned with current and latency-compensated future time steps. 
Other work characterizes these tradeoffs within perception modules, but outside closed-loop control~\cite{huang2017speed}. Crucially, none of these prior works support run-time selection of different resolutions using a single model. While D$^3$~\cite{gog2022d3} explores runtime switching between multiple pre-trained perception models based on environmental conditions, this requires maintaining several models in memory, limiting its applicability in resource-constrained embedded systems.

To address this gap, we leverage dynamic-resolution networks, which—through techniques such as resolution-aware batch normalization—enable a single backbone to efficiently support multiple input resolutions for image classification tasks~\cite{wang2020rsnets,zhu2021drnet}. Building on this principle, we propose a multi-resolution end-to-end model for autonomous driving, specifically tailored to urban environments.

We instantiate a pre-trained, fixed-resolution single-image policy~\cite{chen2021learning} and augment it with a multi-resolution backbone using shared convolutions and per-resolution batch normalization. This design maintains resolution-specific statistics within a single network, enabling operation at multiple input resolutions and facilitating runtime scale selection under a dynamic latency budget. Furthermore, we show that this pre-trained policy can be efficiently retargeted to new input sizes—without access to the original training data—via backbone-only fine-tuning while preserving the driving control head.

We make the following contributions:
\begin{itemize}
\item We empirically study the joint latency–accuracy–safety tradeoff for urban E2E driving, highlighting environment-dependent needs (e.g., intersections vs.\ lane following).
\item We present a single-model, multi-resolution E2E architecture that supports multiple input scales and allows runtime scale selection without duplicating networks.
\item We demonstrate resolution retargeting to new input sizes without access to the original training set, preserving policy behavior while enabling operation under different latency budgets.
\end{itemize}

These results collectively show that resolution adaptivity is a practical lever for improving the latency–safety frontier relative to fixed-resolution baselines.

%% file: 02-background.tex
\section{Background and Motivation}

% In this section, we provide necessary background. 
%on end-to-end autonomous vehicles, latency-accuracy trade-offs and compute-latency tradeoff. 

\subsubsection{End-to-end Deep Neural Networks for Autonomous Vehicles}

% from DeepPicarMicro
% Self-driving cars have been a topic of increasing interest over the past several years. 
% A standard approach is to split the task into multiple specialized sub tasks, such as planning and perception~\cite{autoware}.
Unlike standard robotics control approaches that decompose the task into multiple subtasks—each responsible for a specific, predefined function—the end-to-end (E2E) deep learning approach uses a single deep neural network to produce control outputs directly from raw sensor data. This dramatically simplifies the control pipeline~\cite{levine2016end} and has been explored for decades~\cite{pomerleau1988alvinn,bojarski2016end,bechtel2018deeppicar,bechtel2022deeppicarmicro}.

\begin{figure}[htp]
  \centering
  \includegraphics[width=\linewidth]{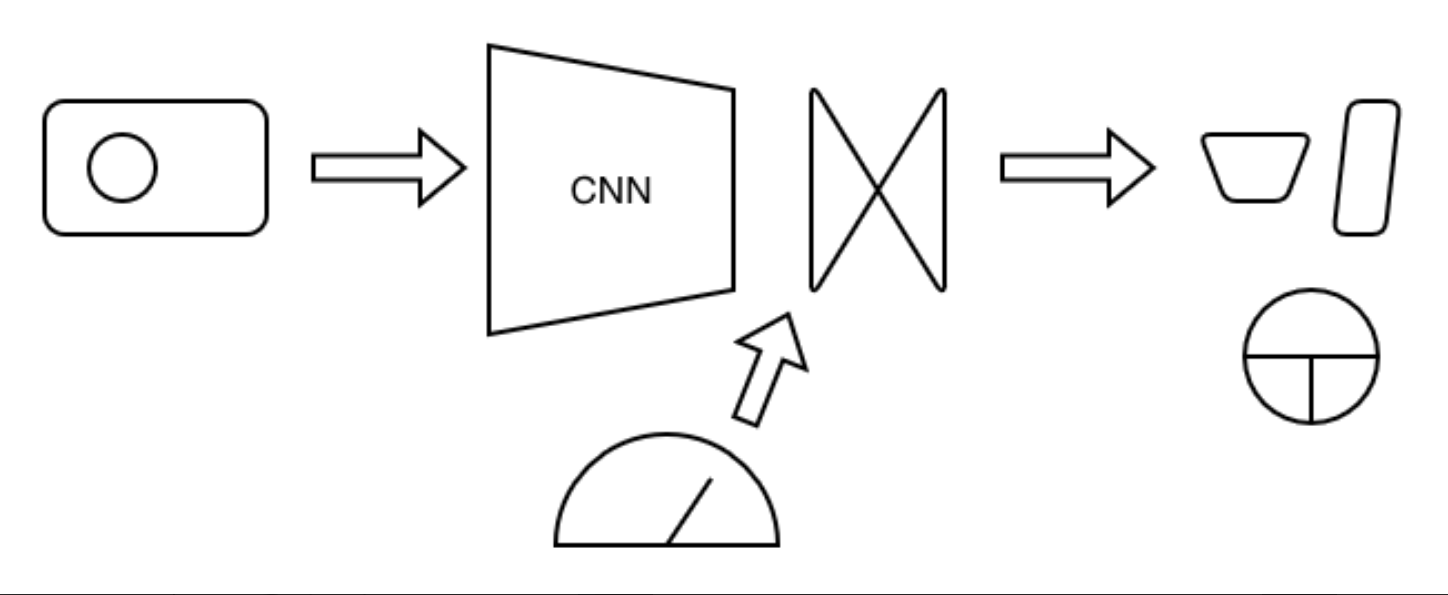}
  \caption{End-to-end policy overview. A CNN backbone encodes the camera stream; ego-speed is provided as a lightweight proprioceptive input; a policy head (MLP) produces vehicle control commands.}
  \label{fig:e2e_overview}
\end{figure}

Fig.~\ref{fig:e2e_overview} shows a high-level overview of a typical end-to-end system, composed of sensors, a DNN, and control actuators.

% Option A: include a pre-made PDF/PNG/SVG figure
% Figure (generic, model-agnostic)

\subsubsection{Importance of Latency in Autonomous Driving}
In an E2E autonomous driving system, 
%camera-to-control driving is a cyber-physical control loop: 
a software policy, a DNN, observes the system status via sensors and directly issues low-level steering/throttle/brake commands. 
Safety of such a system depends on \emph{accuracy}---the ability to perceive and react to small, critical cues (traffic lights, lane markings, curbs, pedestrians)---and on \emph{latency}, the delay $L$ between sensing and actuation. 

This is because each command at time $t$ is computed from measurements taken $L$ seconds earlier, the vehicle has moved $vL$ meters by the time the command takes effect (at speed $v$), so the controller operates on stale information. Furthermore, the action is computed for an earlier world state and applied to a later one, introducing input-output phase lag and a state mismatch (approximately $\Delta s \!\approx\! vL$, $\Delta \theta \!\approx\! \omega L$ under a kinematic model with speed $v$ and yaw rate $\omega$). 
This latency-induced displacement also contributes a term $vL$ to the stopping distance (with maximum deceleration $a_{\max}$)~\cite{falanga2019fast}, % how fast is too fast paper by Scaramuzza
\[
d_{\text{stop}} \;=\; \frac{v^2}{2a_{\max}} \;+\; vL \, .
\]
As $L$ grows or fluctuates, stability margins shrink and tracking errors grow. 
Optimizing accuracy alone is therefore insufficient; \textit{latency is a primary design constraint} that must be \emph{specified, measured, and enforced}. Thus, accuracy and latency jointly determine closed-loop safety.

\subsubsection{Resolution and Latency Tradeoff.}
Convolutional neural networks (CNNs) apply local filters over the spatial field; the dominant compute and memory costs scale with input area ($\Theta(H\cdot W)$ for fixed channel widths). Higher input resolution typically improves recognition of small objects and fine structure but increases %$L_{\text{infer}}$ 
inference latency and memory pressure. Conversely, lower resolution reduces inference latency but risks losing critical cues at urban intersections. Thus, input resolution directly trades perceptual fidelity for end-to-end responsiveness.

\subsubsection{Implications}
Urban routes combine relatively easy \emph{lane-following} segments, where coarse detail often suffices, with \emph{intersections}, where small-object perception is essential. Lane-following is also typically executed at higher speeds, so a given latency $L$ induces a larger state displacement $\Delta s = vL$ (and angular error $\Delta \theta \approx \omega L$) and a larger contribution $vL$ to the stopping distance $d_{\text{stop}}$. Because the latency--accuracy operating point that maximizes safety depends on both scene context and available compute, fixed-resolution policies are inherently suboptimal under changing conditions. This motivates architectures and training protocols that expose resolution as a controllable degree of freedom while respecting closed-loop latency budgets.

%% file: 03-multires.tex
\section{Multi-Resolution End-to-End Deep Neural Network for Autonomous Driving}
\label{sec:method}

\begin{figure}[ht]
    \centering
    \includegraphics[width=0.9\linewidth]{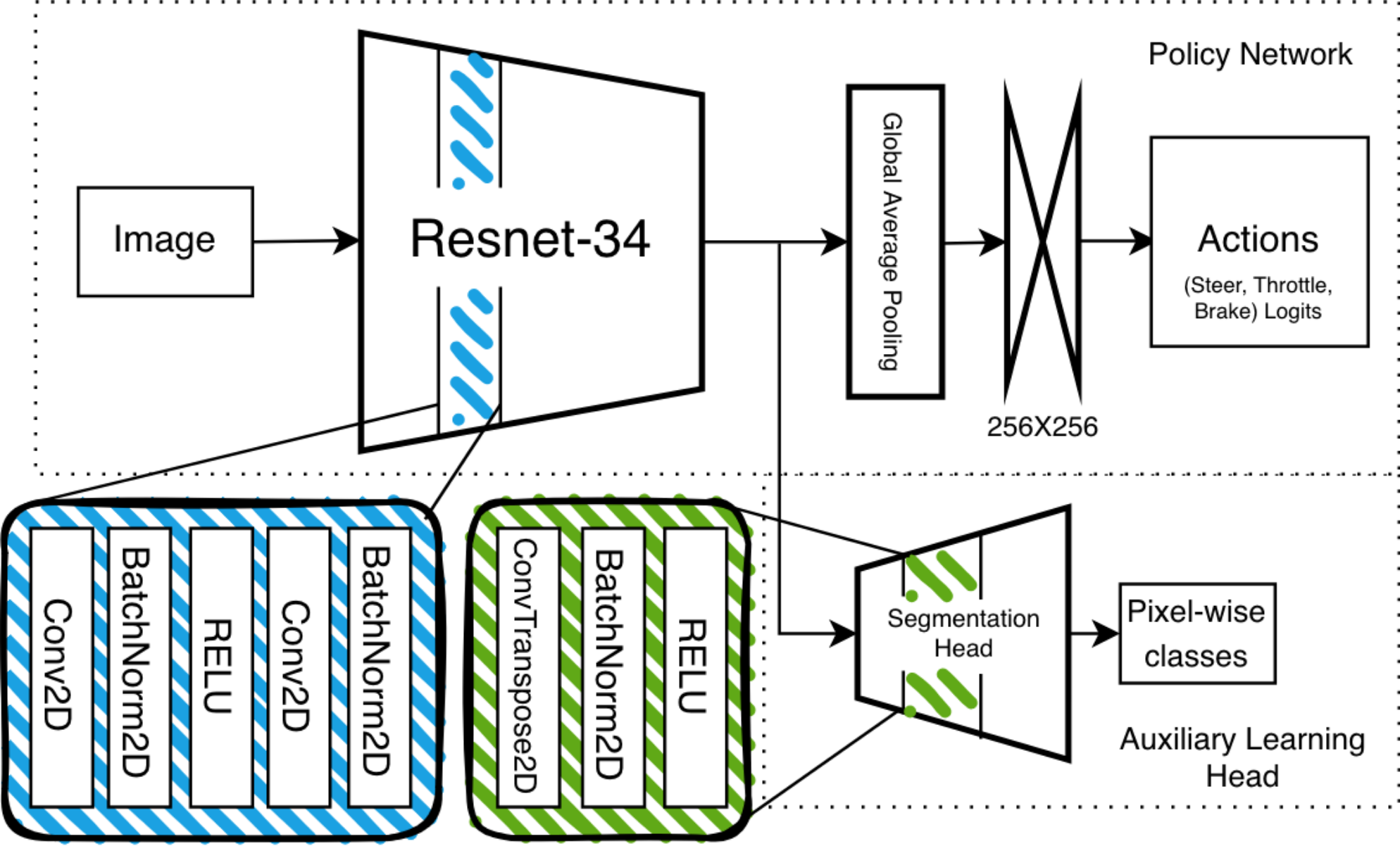}
    \caption{Baseline network architecture }
    \label{fig:architecture}
\end{figure}

% \subsection{Base Policy}
We adopt a monocular camera–based E2E controller from \emph{Learning to Drive from a World on Rails  (WoR)}~\cite{chen2021learning}, which was trained on its \textit{NoCrash} dataset.
%and initialized from the public single-image checkpoint. 
The network uses a ResNet-34~\cite{he2016deep} backbone (ImageNet-pretrained) to encode the camera stream. 
Figure~\ref{fig:architecture} shows its overall architecture.
%and its policy head follows the “speed augmentation” (SA) design: it predicts action-values over discretized speed bins with test-time interpolation at the current ego-speed and uses the original camera augmentations.

%Route-command conditioning matches WoR (lane follow / left / right / straight). Inputs are normalized with ImageNet statistics (mean = [0.485, 0.456, 0.406], std = [0.229, 0.224, 0.225]). In all experiments, the SA mechanism and MLP head are fixed; 
Our only modification to the policy network is to the convolutional backbone (ResNet-34), which we extend to support multiple input scales via per-resolution batch normalization, as described below. Note that the segmentation head is also extended to support multiple input scales, but it is only used during training for auxiliary learning as in the original setup.

\subsection{Per-Resolution Batch Normalization}
\label{sec:multires_backbone}
We expose input scale as a controllable degree of freedom by sharing all convolutional weights and replicating \emph{batch normalization} (BN) per resolution. Let $\mathcal{R}=\{r_1,\dots,r_m\}$ be the set of aspect-ratio–preserving scale factors relative to the WoR input size $S_0=(H_0,W_0)$. For each backbone BN layer $\mathrm{BN}_\ell$ with $C_\ell$ channels we create a bank
\[
\big\{\mathrm{BN}_\ell^{(r)} : r\in\mathcal{R}\big\},\qquad
\mathrm{BN}_\ell^{(r)} \equiv \big(\gamma_\ell^{(r)},\beta_\ell^{(r)},\mu_\ell^{(r)},\sigma_\ell^{(r)}\big),
\]
while keeping all convolutional weights $\{W_\ell\}$ shared. Initialization copies the pretrained BN affine parameters and running statistics into every branch, $\mathrm{BN}_\ell^{(r)} \leftarrow \mathrm{BN}_\ell$.

%\textbf{Inference and overhead.} 
At test time, exactly one BN branch per layer is active, matching the chosen input scale $r$; this introduces no extra FLOPs beyond those implied by the input size. Parameter overhead is limited to the BN banks (per-scale $\gamma,\beta$ and running $\mu,\sigma$) and is negligible relative to convolutions.

\subsubsection{Multi-resolution training step}
At each optimizer step we process \emph{all} scales in $\mathcal{R}$ \emph{sequentially}. For every $r\in\mathcal{R}$: (i) resize the entire micro-batch to $(\lfloor rH_0\rfloor,\lfloor rW_0\rfloor)$ (no mixing of scales within a micro-batch), (ii) activate only the corresponding per-resolution batch-normalization branches $\{\mathrm{BN}_\ell^{(r)}\}$, (iii) run a forward pass and accumulate that scale’s losses. We then average across $r\in\mathcal{R}$ and perform a \emph{single} backward/optimizer step, updating the shared convolutions and the active BN branch for each scale; the policy head remains frozen.

\subsection{Resolution Retargeting}
\label{subsec:resolution_retarget}%  without Original Data}

\subsubsection{Data availability and provenance}
WoR~\cite{chen2021learning} does not 
%release a large CARLA Leaderboard dataset utilized for training a model architecture using a multi-camera rig but do not 
release the single-image \textit{NoCrash} training set~\cite{chen2021learning} used to train the baseline model. Therefore, we initialize from the WoR's public single-image checkpoint and perform resolution retargeting using non-overlapping urban routes collected separately with the same sensor configuration and augmentation protocol.

\subsubsection{Protocol} We adapt the CNN backbone to new input sizes while keeping the policy head 
%(including SA) 
fixed. We choose $\mathcal{R}=\{0.75,\,1.0\}$ as aspect-ratio–preserving scale factors relative to the WoR input size $S_0$. Training follows the multi-resolution schedule described above,
with progressive unfreezing ($\mathrm{layer4}\rightarrow \mathrm{layer1}$). %---that is, at each optimizer step we process all $r\in\mathcal{R}$ sequentially, average the scale-specific losses, and perform a single optimizer update that modifies the shared convolutions and the corresponding BN branches; the policy head remains frozen.

\subsubsection{Optimization objective}
Let $x$ denote the input image and $c$ the route command. For a scale factor $r\in\mathcal{R}$, the network produces action-value logits $z^{(r)}(x,c)\in\mathbb{R}^{|\mathcal{A}|}$ over a discrete action set $\mathcal{A}$. Following WoR~\cite{chen2021learning}, we form a target action distribution by applying a softmax (“Boltzmann policy”) to action-values supplied in the released dataset as ground-truth; denote this target by $t(\cdot\mid x,c)$ with temperature $\tau_0$. The base supervision at scale $r$ is the KL divergence from the target to the policy:
\[
L_{\mathrm{act}}^{(r)}
=
D_{\mathrm{KL}}\!\Big(
t(\cdot\mid x,c)
\,\Big\|\,
\mathrm{softmax}\!\big(z^{(r)}(x,c)\big)
\Big).
\]

\subsubsection{Auxiliary segmentation}
Following WoR~\cite{chen2021learning}, we include semantic segmentation as an auxiliary objective and \emph{train} the segmentation decoders jointly with the backbone. All batch normalization layers—including those in the segmentation branches—use per-resolution banks to avoid cross-scale statistic mismatch. The decoder is fully convolutional and operates on spatial maps without global pooling, so batch-normalization moments
$\mu_c = \mathbb{E}_{b,h,w}[x_{b,c,h,w}]$ and
$\sigma_c^2 = \mathrm{Var}_{b,h,w}[x_{b,c,h,w}]$
vary with input scale. For scale $r$, let $s^{(r)}$ be the logits and $y$ the ground-truth mask at label resolution. We resize logits to the label size (nearest neighbor) and use pixelwise cross-entropy:
\[
L_{\text{seg}}^{(r)}=\mathrm{CE}\!\big(\mathrm{upsample}(s^{(r)}),\,y\big).
\]

\subsubsection{Full-resolution teacher regularization}
To preserve the original policy while adapting to new input sizes, we distill from a \emph{frozen} copy of the public checkpoint at full resolution ($1.0\times$) as a teacher model. Let $z^{T}(x,c)$ and $f^{T}(x)$ be the teacher logits and pooled backbone features with dimensionality $D$, and $f^{(r)}(x)$ the student features at scale $r$. We use:

\[
\begin{aligned}
L_{\mathrm{KD}}^{(r)} &=
D_{\mathrm{KL}}\!\big(\operatorname{softmax}(z^{T}) \,\big\|\, \operatorname{softmax}(z^{(r)})\big),\\
L_{\mathrm{feat}}^{(r)} &= \frac{1}{D}\,\big\|f^{(r)}(x) - f^{T}(x)\big\|_2^2.
\end{aligned}
\]

\subsubsection{Aggregate retargeting loss}
Averaging over scales yields
\[
L_{\mathrm{retarget}}
=
\frac{1}{|\mathcal{R}|}\sum_{r\in\mathcal{R}}
\Big(
L_{\mathrm{act}}^{(r)}
+\lambda_{\mathrm{seg}}\,L_{\mathrm{seg}}^{(r)}
+\lambda_{\mathrm{KD}}\,L_{\mathrm{KD}}^{(r)}
+\lambda_{\mathrm{feat}}\,L_{\mathrm{feat}}^{(r)}
\Big),
\]
with $\lambda_{\mathrm{seg}},\lambda_{\mathrm{KD}},\lambda_{\mathrm{feat}}$ set per experiment. The teacher is stop-gradient; student and teacher use identical data augmentations, differing only by the student’s isotropic resize to scale $r$.

%% file: 04-evaluation.tex
\section{Evaluation}

In this section, we present our evaluation setup and the results. 

\subsection{Experimental Setup}
\subsubsection{Simulator and timing}
All experiments run in the CARLA synchronous mode with the simulation tick at 40\,Hz and camera sensors at 20\,Hz.
We evaluate three settings: (a) 50\,ms period, no injected delay ($D{=}0$); (b) 50\,ms period with injected delay ($D{=}50$\,ms); and (c) latency-matched runs where we set $D\in\{100,150,200\}$\,ms and choose the controller period $T_c{=}D$. Injected delay is realized via zero-order hold: after a new command is computed, the previously applied command is held for $D$ before the new command is latched. In synchronous CARLA, the achieved loop period is $T_{\text{loop}}=\max(T_c,\,L_{\text{exec}})$. When $L_{\text{exec}}\le T_c$, the measured sensing-to-actuation delay is $\approx D$; if $L_{\text{exec}}>T_c$, it is $\approx L_{\text{exec}}$ (compute-bound).

\subsubsection{Routes and splits}
We evaluate on the two CARLA towns used by the WoR \textit{NoCrash} policy: Town01 (seen during WoR training) and Town02 (unseen test). We reuse the PCLA codebase~\cite{tehrani2025pcla} to load the WoR single-image checkpoint as a CARLA agent and configure sensors and routes. Each town contributes \emph{two} routes (four routes total). Unless noted, per-configuration summaries for a given town aggregate over its two routes and three traffic densities.

\subsubsection{Traffic workloads and trials}
Traffic is spawned by CARLA’s navigation mesh with randomized seeds, so realized counts can deviate from targets. We test three target densities \((\#\mathrm{veh},\#\mathrm{ped})\in\{(0,0),(5,10),(20,40)\}\). To characterize this variability, for each (route, density) pair we perform 10 independent environment resets and record the realized vehicle/pedestrian counts; Table~\ref{tab:traffic_workloads_combined} reports the resulting mean$\pm$sd per route and town.
For the main evaluation, we then run one rollout per (route, density) pair—two routes $\times$ three densities per town—yielding \(6\) runs per town (12 total per configuration), and report performance metrics aggregated over these runs.

% Replace the previous combined table with this version.
\begin{table}[t]
  \centering
  \footnotesize
  \setlength{\tabcolsep}{2pt}
  \caption{Simulated traffic conditions 
  %(mean$\pm$sd over 10 independent environment resets per (route, density))
  using CARLA's randomized spawner.}
  \label{tab:traffic_workloads_combined}
  \begin{tabular}{l cc|cc|cc|cc}
    \toprule
      & \multicolumn{4}{c|}{Town01} & \multicolumn{4}{c}{Town02} \\
    \cmidrule(lr){2-5}\cmidrule(lr){6-9}
    \multirow{2}{*}{\shortstack{Target\\(veh, ped)}}
      & \multicolumn{2}{c}{Route 1} & \multicolumn{2}{c|}{Route 2}
      & \multicolumn{2}{c}{Route 1} & \multicolumn{2}{c}{Route 2} \\
      & Veh & Ped & Veh & Ped & Veh & Ped & Veh & Ped \\
    \midrule
    (0,\;0)   & 0  & 0             & 0  & 0             & 0  & 0             & 0  & 0 \\
    (5,\;10)  & 5  & $7.0\!\pm\!1.5$ & 5  & $7.0\!\pm\!1.5$ & 5  & $6.6\!\pm\!1.4$ & 5  & $6.6\!\pm\!1.4$ \\
    (20,\;40) & 20 & $26.7\!\pm\!3.1$& 20 & $26.5\!\pm\!3.1$& 20 & $27.2\!\pm\!2.9$& 20 & $27.2\!\pm\!2.9$ \\
    \bottomrule
  \end{tabular}
\end{table}

\subsection{Effects of Resolution Retargeting}
In this experiment, we investigate the impact of resolution retargeting on driving performance of the evaluated models.

Table~\ref{tab:non_degradation} reports success (route completion) rates and recoverable collision incidences (over 6 runs per town) of the Original WoR policy (Original), two fixed-resolution baselines (NRA 0.75×/1.0×), and our multi-resolution policy at 0.75×/1.0× (RA). The NRA baselines are models without BN banks for each resolution, but fine-tuned using the same method described in Subsection~\ref{subsec:resolution_retarget}. The number denotes the image-resolution scaling factor. 
Both NRA and RA configurations are resolution retargeted---i.e., fine-tuned for target resolutions on 50k frames from the 1M-frame CARLA Leaderboard dataset. Originally, all models were pre-trained on the 270k Town01 \textit{NoCrash} dataset.

%Note that for RA, these results come from the same model weights applied at different input resolutions, whereas the NRA models are separately trained for each resolution.

On Town01, all models complete the route without any collisions. On Town02, all models also successfully complete the route but resolution retargeted models (NRA and RA) experience 1 or 2 collisions, which they can recover to complete the route. From this, we find that resolution retargeting minimally impact performance despite being fine-tuned with a different dataset. Furthermore, we find that the two RA configurations with a single shared model perform similarly with the NRA baselines. 
%In other words, our single multi-resolution model, operating at two different resolutions, matches performance of the fixed resolution baselines under the same training condition. 
In the following, we focus on comparing the performance of NRA and RA in more details.
%success is also 6/6; collision incidence is 1/6 for NRA~0.75$\times$, NRA~1.0$\times$, and RA~0.75$\times$, and 2/6 for RA~1.0$\times$. Although the Original WoR policy had 0/6 collisions in our Town02 batch, WoR’s NoCrash table~\cite{chen2021learning} reports sub-100\% success on the test town, indicating occasional collisions. With $n{=}6$ per town, such one–two-event differences fall within binomial sampling noise, so the configurations are comparable; 
%RA matches NRA in success and collision incidence, despite both being fine-tuned on 50k frames from the 1M-frame CARLA Leaderboard set using the dataset’s wide image, rather than WoR’s 270k Town01 \textit{NoCrash} set.

\begin{table}[t]
  \centering
  \footnotesize
  \setlength{\tabcolsep}{3pt}
  \caption{Driving performance comparison. NRA: non--resolution-aware; RA: resolution-aware.}
  \label{tab:non_degradation}
  \begin{tabular}{lccccc}
    \toprule
    Map & Orig. & NRA 0.75$\times$ & NRA 1.0$\times$ & RA 0.75$\times$ & RA 1.0$\times$ \\
    \midrule
    \multicolumn{6}{l}{\emph{Success (\%)}} \\
    Town01 & 100 & 100 & 100 & 100 & 100 \\
    Town02 & 100 & 100 & 100 & 100 & 100 \\
    \midrule
    \multicolumn{6}{l}{\emph{Collision incidence}} \\
    Town01 & 0 & 0 & 0 & 0 & 0 \\
    Town02 & 0 & 1 & 1 & 1 & 2 \\
    \bottomrule
  \end{tabular}
\end{table}

\subsection{Effects of Input Resolution}
In this experiment, we investigate the impact of input resolution on the performance of the evaluated models.

\begin{figure}[t]
  \centering
  \includegraphics[width=0.92\columnwidth]{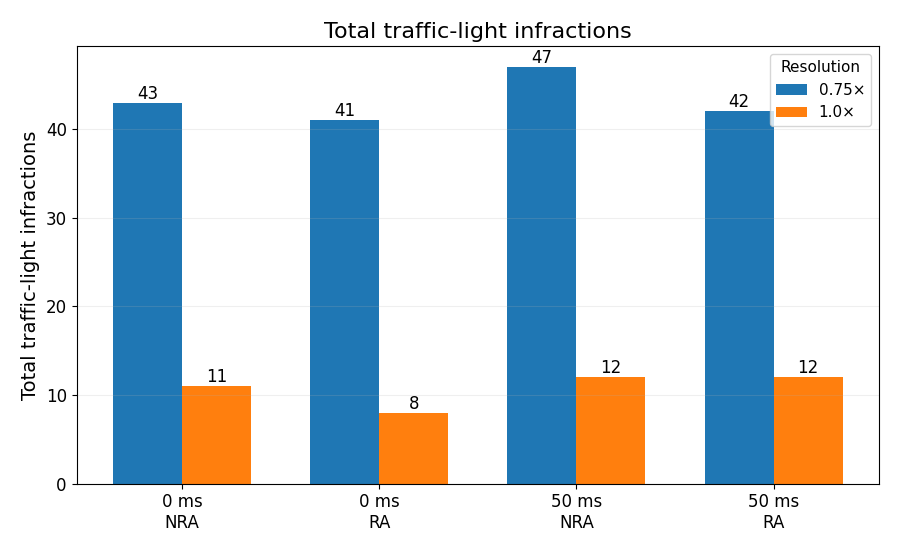}
    \caption{Traffic-light infractions vs.\ input scale at a 50\,ms control period, with/without 50\,ms injected delay.}
    %Downscaling strongly increases infractions; RA tracks NRA.}
  \label{fig:latency_infractions}
\end{figure}

Fig.~\ref{fig:latency_infractions} shows the traffic-light infraction rates of the RA and NRA models under varying latency. As illustrated in the figure, when the control period is fixed at 50 ms, reducing the input scale from the training-aligned $1.0\times$ to $0.75\times$ significantly increases traffic-light infractions across all configurations.

\begin{figure}[t]
  \centering
  \includegraphics[width=0.92\columnwidth]{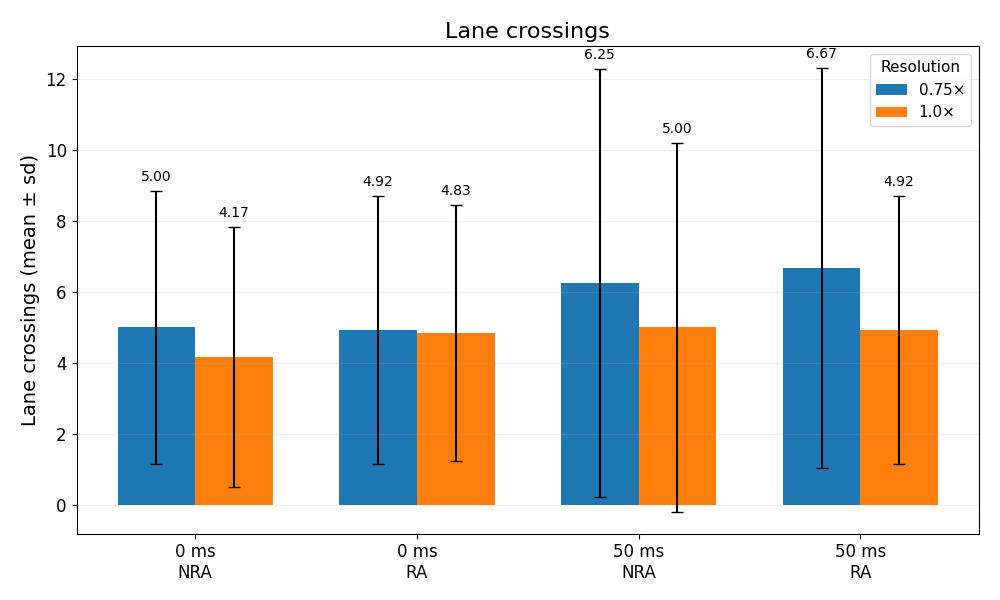}
  \caption{Lane invasions per route (mean$\pm$sd) at a 50\,ms control period, with/without 50\,ms injected delay. } 
  % Downscaling to $0.75\times$ mildly increases invasions; RA mirrors NRA.}
  \label{fig:latency_lanes}
\end{figure}

Fig.~\ref{fig:latency_lanes}, on the other hand, shows the lane-invasion rates. Unlike traffic-light infractions, lane-invasion rates do not vary significantly with lower resolution. This is because traffic-light recognition relies on high resolution due to the small size of the lights, whereas lane detection is far less sensitive to reduced resolution.

These results show that input resolution materially influences urban E2E driving quality—particularly compliance with small-object cues—independent of the model variant. 
%This aligns with our hypothesis that resolution is a primary lever in the latency–accuracy tradeoff.

\subsection{Effects of Latency}
In this experiment, we investigate the impact of a model's execution time (simulated by added latency) to its driving performance. 

\begin{figure*}[t]
  \centering
  \subfloat[NRA (non--resolution-aware)\label{fig:nra_latency}]{
    \includegraphics[width=0.48\textwidth]{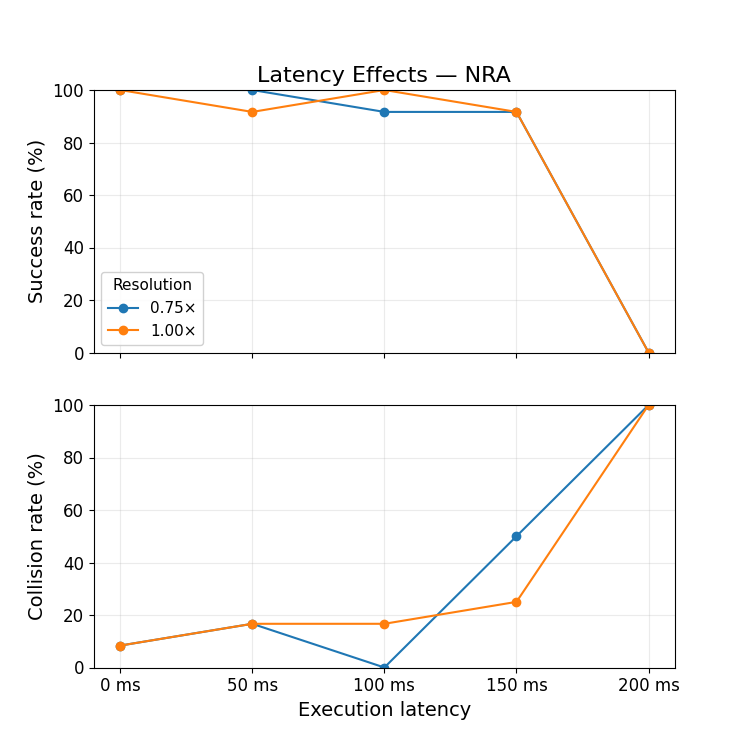}
  }\hfill
  \subfloat[RA (resolution-aware)\label{fig:ra_latency}]{
    \includegraphics[width=0.48\textwidth]{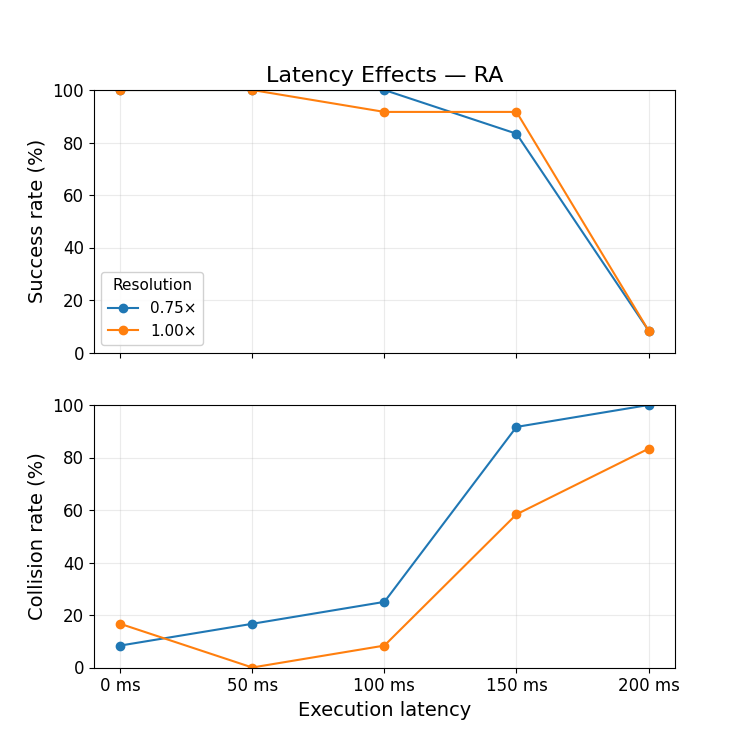}
  }
  \caption{Success and collision rates vs.\ injected delay ($D\in\{0,50,100,150,200\}$\,ms) and input scale ($0.75\times,1.0\times$). Percentages are over 12 runs/configuration.}
  \label{fig:latency_res_effects}
\end{figure*}

Fig.~\ref{fig:latency_res_effects} shows the results. 
Across both NRA and RA models and at both scales, increasing injected delay produces a sharp transition from near-perfect operation to frequent catastrophic failure. With $D\!\le\!100$\,ms, success (route completion) rates remains high (91.65–100\%) and collision rates are modest (0–25\%). At $D\!=\!150$\,ms, however, collision incidence jumps
%( NRA~$0.75\times$: 50\%; RA~$0.75\times$: 91.65\%; RA~$1.0\times$: 58.35\%) 
and success begins to drop significantly. % (83.35–91.65\%). 
By $D\!=\!200$\,ms, success collapses to $0$–$8.35$\% and collisions reach $83$–$100$\% across configurations. 

These results indicate that safety degrades as latency increases; moreover, the degradation is non-linear, with a ``cliff'' around $\sim\!150$\,ms where success collapses and collisions spike. 
%Percentages change in $\approx8.35\%$ steps because each configuration aggregates 12 runs.

\subsection{Runtime Latency-Accuracy Tradeoffs}

In this experiment, we investigate the potential benefits of our approach by dynamically switching resolutions depending on the environment compared to the fixed resolution baselines. 

%\paragraph{Anytime (oracle) switching improves safety within a fixed latency envelope.}
Here, we assume an ``oracle'', which decides which resolution is ideal for a given environment. That is, it triggers a resolution switch based on ego vehicle proximity to the traffic lights (default $0.75\times$; switch to $1.0\times$ within 5\,m of a traffic light; revert after exiting a 15\,m radius). The intuition is that because higher resolution improve traffic light detection, such a policy can improve driving performance. 

Table~\ref{tab:ra_vs_fixed} shows the results. Note that
the resolution–aware (RA) switcher outperforms both fixed baselines under the \emph{same execution latency}. 
Under the $100{\to}50$\,ms envelope, RA holds 100\% success with 0\% collisions, reduces lane invasions from $6.67\!\pm\!5.63$ (fixed $0.75\times$) to $5.50\!\pm\!5.21$ ($\sim\!17.5\%$ lower), and cuts red-light violations from 42 to 15 ($\sim\!2.8\times$ fewer).
%, while matching the fixed $1.0\times$ red-light count (15 vs.\ 14) but avoiding its lower success (91.65\%) and higher lane errors ($9.84\!\pm\!9.33$). 
Under the tighter $150{\to}100$\,ms envelope, RA again achieves 100\% success and 0\% collisions, lowers lane invasions from $11.25\!\pm\!9.14$ (fixed $0.75\times$) to \textbf{$10.84\!\pm\!9.10$} ($\sim\!3.6\%$ lower), and reduces red lights from 43 to 17 ($\sim\!2.5\times$ fewer), significantly outperforming either of the fixed resolution baselines. 
%the fixed $1.0\times$ alternative at 150\,ms shows degraded safety (58.35\% collisions; $66.5\!\pm\!76.0$ lane crossings; 91.65\% success). 

These results indicate that selectively spending compute near small-object bottlenecks (traffic lights) materially improves the latency–safety operating point compared to any fixed-scale policy with the same latency ceiling. 
%Lane crossing is also improved because under RA can utilizes the lower resolution input in high speed and the high resolution near crossing, which improve performance.

%We implement the trigger as an oracle to isolate the potential of resolution switching; practical deployments can approximate this with map priors or lightweight signal detectors.
Note that the oracle trigger can be practically implemented by leveraging a map information as typically available in actual autonomous driving systems. 

% \paragraph{Why RA yields fewer lane crossings than either fixed baseline.}
% The anytime policy combines the strengths of each fixed setting. Near intersections it switches to $1.0\times$, improving geometric and small-object perception and producing smoother, more decisive turns; the additional latency has limited impact because speeds are low, so the displacement term $vL$ is small. On lane-following stretches it runs at $0.75\times$ with lower inference delay, reducing phase lag and boundary oscillations that the higher-latency $1.0\times$ setting exhibits at higher speeds. Consequently lane invasions fall below both fixed policies under the same latency ceiling (e.g., $100{\to}50$\,ms: $2.26{\pm}0.71$ vs.\ $6.67{\pm}5.63$ and $9.84{\pm}9.33$; $150{\to}100$\,ms: $3.33{\pm}0.25$ vs.\ $11.25{\pm}9.14$ and $66.5{\pm}76.0$; Table~\ref{tab:ra_vs_fixed}).

\begin{table*}[t]
\centering
\caption{Resolution-aware (RA) switching versus fixed-resolution baselines under the same worst-case execution latency $L_{\text{high}}$.}
%Lower is better for collision rate, lane crossings, and red lights.}
\label{tab:ra_vs_fixed}
\setlength{\tabcolsep}{6pt}
\begin{tabular}{llrrrrr}
\toprule
Latency envelope & Configuration & Runs & Success (\%) & Collision (\%) & Mean lane & Red lights \\
$L_{\text{high}}\!\to\!L_{\text{low}}$ &  &  &  &  & crossings & (total) \\
\midrule
\multirow{3}{*}{$100\!\to\!50$ ms}
  & Fixed 0.75$\times$ @ 50\,ms     & 12 & 100.00 & 16.65 & $6.67 \pm 5.63$ & 42 \\
  & \textbf{RA switcher (50/100\,ms)} & \textbf{12} & \textbf{100.00} & \textbf{0.00}  & $\mathbf{5.50 \pm 5.21}$ & 15 \\
  & Fixed 1.0$\times$ @ 100\,ms     & 12 & 91.65  & 8.35  & $9.84 \pm 9.33$ & \textbf{14} \\
\addlinespace
\multirow{3}{*}{$150\!\to\!100$ ms}
  & Fixed 0.75$\times$ @ 100\,ms    & 12 & 100.00 & 25.00 & $11.25 \pm 9.14$ & 43 \\
  & \textbf{RA switcher (100/150\,ms)}& \textbf{12} & \textbf{100.00} & \textbf{0.00}  & $\mathbf{10.84 \pm 9.10}$ & \textbf{17} \\
  & Fixed 1.0$\times$ @ 150\,ms     & 12 & 91.65  & 58.35 & $66.50 \pm 76.01$ & 18 \\
\bottomrule
\end{tabular}
\end{table*}

%% file: 05-related.tex
\section{Related Work}

\subsubsection{System-level latency-aware AV frameworks} A line of work focuses on managing latency and accuracy at the \emph{system} level in modular autonomous driving stacks. Pylot is a modular AV platform built on a dataflow runtime that enables swapping perception and planning components and explicitly studying how their latency and accuracy affect end-to-end driving behavior in CARLA and on real vehicles~\cite{gog2021pylot}. Building on this stack, D$^3$ introduces a dynamic deadline-driven execution model and runtime that centralizes deadline management and adapts the computation graph to changing driving context and resource availability, reducing collisions by adjusting module scheduling and quality under tight timing constraints~\cite{gog2022d3}. These frameworks treat latency as a first-class concern at the pipeline level—deciding which modules to run and at what fidelity—but assume fixed-resolution perception networks trained separately and kept resident in memory at runtime—often impractical on embedded GPUs. 
%and do not endow a single camera-to-control policy with built-in multi-resolution behavior or data-free resolution retargeting as we do.

\subsubsection{Anytime perception modules for autonomous driving}
A complementary line of work develops perception components that trade accuracy for latency at run time, independent of the controller. In vision, \textit{anytime} and budget-aware CNNs refine predictions progressively or use early exits to provide outputs under tight budgets; for example, anytime stereo networks generate intermediate disparity maps to enable latency–accuracy tradeoffs on embedded robots~\cite{wang2019anytime}. For object detection, dynamic and early-exit designs such as  AdaDet~\cite{yang2023adadet} and AnytimeYOLO~\cite{kuhse2025you} adjust computation per frame while maintaining detection quality under inference-time limits. Others explored anytime processing of 3D LiDAR point clouds~\cite{soyyigit2025mural,soyyigit2022anytime,soyyigit2024valo}.
In particular, MURAL supports multi-resolution inference to meet per-frame deadlines~\cite{soyyigit2025mural}.
These approaches demonstrate the value of deadline-aware, multi-resolution, or early-exit perception, but they remain within modular stacks and are evaluated on perception metrics, not closed-loop behavior. In contrast, we integrate multi-resolution capability directly into an end-to-end driving policy and evaluate its effect on urban driving safety under latency constraints.

%% file: 06-conclusion.tex
\section{Conclusion}
In this work, we studied how input resolution shapes the latency–accuracy trade space in closed-loop, end-to-end (E2E) urban driving. Building on a WoR single-image policy, we introduced a multi-resolution backbone with per-resolution batch normalization that (i) preserves baseline behavior, (ii) enables runtime selection of input scale under a latency budget, and (iii) supports resolution retargeting via backbone-only fine-tuning.

Two findings emerge across CARLA routes and traffic workloads. First, resolution strongly influences safety-critical behavior at a fixed control period: downscaling from the training-aligned size increases red-light violations and lane invasions, showing that small-object cues and geometric fidelity matter even when loop timing is constant. Second, latency shows a nonlinear “failure cliff”: as injected delay nears $\sim150$,ms, success collapses and collisions spike for all variants. Within these limits, we showed that a simple policy that switches to high resolution near intersections improves the latency–safety frontier relative to any fixed scale at the same latency ceiling.

Our resolution-aware backbone is lightweight (only BN banks are replicated), leaves the control head unchanged, and supports retargeting to new input sizes using non-overlapping routes, making resolution a practical runtime knob under compute constraints.

\subsubsection{Limitations and future work}
Our switching trigger is oracle-based and tested only in two simulated towns; broader generalization, hardware-in-the-loop timing, and jitter robustness remain open. Future directions include learned or map-based triggers, integration with explicit latency controllers, and real-vehicle evaluation under strict latency accounting.

%% file: 07-ack.tex
\section*{Acknowledgment}
\label{sec:acknowledge}

This research is supported in part by NSF CPS-2038923 and 
Department of Defense (DoD) Contract number HQ00342310007.